# Named Entity Recognition in Electronic Health Records Using Transfer Learning Bootstrapped Neural Networks

**Luka Gligic**
University of Oxford
luka.gligic@gtc.ox.ac.uk

**Andrey Kormilitzin**
University of Oxford
andrey.kormilitzin@psych.ox.ac.uk

**Paul Goldberg**
University of Oxford
paul.goldberg@cs.ox.ac.uk

**Alejo Nevado-Holgado**
University of Oxford
alejo.nevado-holgado@psych.ox.ac.uk

## Abstract

Neural networks (NNs) have become the state of the art in many machine learning applications, such as image, sound [1] and natural language processing [2,3]. However, the success of NNs remains dependent on the availability of large labelled datasets, such as in the case of electronic health records (EHRs). With scarce data, NNs are unlikely to be able to extract this hidden information with practical accuracy. In this study, we develop an approach that solves these problems for named entity recognition, obtaining 94.6 F1 score in I2B2 2009 Medical Extraction Challenge [6], 4.3 above the architecture that won the competition. To achieve this, we bootstrap our NN models through transfer learning by pretraining word embeddings on a secondary task performed on a large pool of unannotated EHRs and using the output embeddings as a foundation of a range of NN architectures. Beyond the official I2B2 challenge, we further achieve 82.4 F1 on extracting relationships between medical terms using attention-based seq2seq models bootstrapped in the same manner.

## 1 Introduction

Electronic Health Records (EHRs) are the databases used by hospital and general practitioners to daily log all the information they record from patients [7]. This information typically includes, but is not limited to: disorders, taken medications, dosages, symptoms, results from medical tests, and even considerations made by the doctor when evaluating each patient. In number of subjects (for example, 50 million patients in the case of European Medical Information Framework (EMIF)), EHRs are the largest source of empirical data in biomedical research [4,8], making them ideal for studying disease (e.g. Alzheimer's [9], cardiovascular disease [10], or associated risk factors [11–13]) and evaluating service (e.g. monitoring adverse drug reactions [14]). However, most of the information held in EHRs is in the form of natural language text (i.e. written by the physician during each session with each patient), making it inaccessible for research [4,5]. Unlocking all this information would represent a considerable contribution to biomedical research, multiplying the quantity and variety of scientifically usable data, which is the reason why major efforts have been relatively recently initiated towards this goal [4,8,11,15] as well as being the main motivation behind this work.

The central idea of the paper is to develop an accurate and robust neural model for information extraction from medical texts, specifically, we were interested in medical named entity recognition (NER) and relation extraction (RE) between them.

| Year | Existing annotations | Total documents | Unique documents (not annotated) | Unique documents (annotated) |
|---|---|---|---|---|
| 2007 | Smoking | 2886 | 926 | 0 |
| 2008 | Obesity | 1267 | 1237 | 0 |
| 2009 | Medications | 1945 | 991 | 258 |
| 2010 | Term relations | 696 | 694 | 0 |
| 2011 | Conference | 424 | 188 | 0 |
| 2012 | Temporal relations | 671 | 311 | 0 |
| | **Total** | 7889 | **4347** | **258** |

***Table 1. I2B2 datasets used in this study.*** *Third column indicates the total number of documents in each corpus. Fourth and fifth columns indicate which not annotated and annotated documents, respectively, were unique, and therefore added into the common pool of documents used for subsequent analyses and unsupervised and supervised training.*

Although traditional Natural Language Processing (NLP) algorithms, such as rule systems [16], can perform this task with fair accuracy in the simpler situations (well-structured text, large amounts of labelled data available and many annotated samples), the challenge remains an unsolved problem in the more complex cases (badly structured language, few labelled samples) [17]. Unfortunately, data found in EHRs falls under the second category. Namely, physicians tend to use badly formatted shorthand and non-widespread acronyms ('transport pt to OT tid via W/C' for 'transport patient to occupational therapy three times a day via wheel chair'), while labelled records are scarce (ranging in the hundreds for a given task and with very few annotated samples). A reason for this scarcity is that data access is difficult due to ethical concerns [18–20]. Other reason is that, even with data access granted, medical text needs to be annotated by field expert (e.g. clinicians), who are themselves in short supply.

In the study presented in this paper we address these problems by: first, using Neural Networks (NN) [1,3], which are expected to be more robust to badly structured language than rules or other traditional techniques [2]; second, rather than training them only on the objective task, we bootstrap the Neural Networks through transfer learning, by feeding them pretrained word embeddings from a secondary task on unannotated electronic records. This approach achieves 94.7 F1 in I2B2 2009 Medical Information Extraction challenge, 4.3 more than the traditional approach that originally won the challenge. In addition to the official objectives of I2B2 2009, this approach also obtained 82.4 F1 on extracting the relationships between medical terms, which are of high importance in research with EHRs.

## 2 Methods:

### 2.1. Objective task

Our objective task consisted on automatically locating and predicting the annotations of I2B2 2009 Medical Information Extraction challenge [6]. These labels consisted on all mentions of medications where the patient was the user, plus a number of associated fields per term. These fields were: medication, dosage, mode, frequency, duration, reason. Medication includes compound name, brand name, generics, collectives and prescriptions (e.g. acetylsalicylic acid or aspirin). Dosage indicates the amount administered to the patient, which could be a measurement (e.g. 2.0 mgs) or units (e.g. 2 tablets). Mode refers to the administration route (e.g. orally). Frequency refers to how often the medication was taken (e.g. 2 per day). Duration consists on treatment length (e.g. until symptoms disappear). Reason is the cause for the prescription (e.g. presumed pneumonia).

### 2.2. Datasets

This study used all datasets released by I2B2 from 2007 to 2012. We observed that some documents were repeated across different yearly releases. To eliminate duplicates, we sequentially pooled each corpus into a final set of 4605 unique documents (see Table 1). I2B2 2009 challenge released a total of 1249 unique documents, with 258 of them annotated for the objective task. Given that our objective task was the one corresponding to I2B2 2009 challenge, only the 258 documents from this year were considered annotated for our case, using all others as unannotated samples for the purpose of transfer learning. In detail, 4347 unannotated samples were selected for training embeddings, 238 for training the rest of the NN, 10 for validation and 10 for final testing.

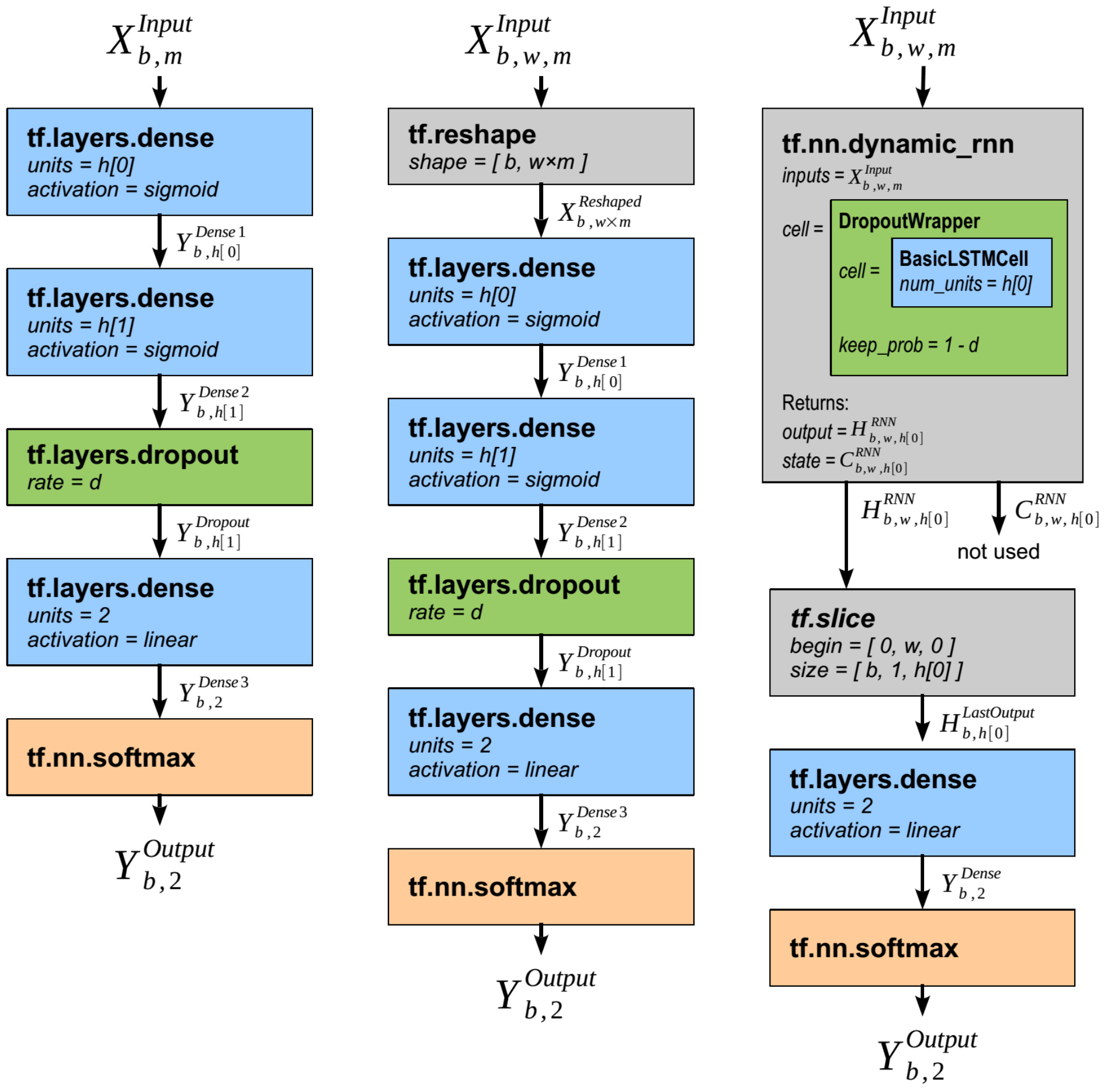

***Figure 1. Architectures for term classification.*** *From left to right, the figure shows the context free FNN, the context aware FNN and the RNN architectures used for terms classification. The component operations (e.g. layers) of each architecture are represented as boxes, with blue for full layers, green for dropout, orange for transformation functions, and grey for shuffling or tensorflow wrappers. Within each box, bold font shows the name of the tensorflow operation, and italic fonts the input parameters when non default values were used for that particular opreation. In occasions, input and output tensors are also represented with a capital letter, with subindex for tensor dimensions, and superindex for a further description of the data held in that particular tensor.*

## 2.3. Text pre/processing

Text was pre-processed to reduce the number of out of vocabulary (OOV) words, which was defined as words not accounted by the embeddings described in section 2.4. Sentences were split on "." followed by a capital letter, as recommended by Patrick and Li [21]. All numbers were replaced by the special token <num>. Punctuation symbols ".", ":" and ";" were removed, unless they were surrounded by letters or followed a number. All letters were lower cased. Pre-processing did not alter number and location of words and sentences. Finally, a number of metrics evaluated the text demographics of the embedding/train/validation/test datasets after pre-processing.

## 2.4. Training embeddings

We created two embeddings versions with Contiguous Bag of Words (CBOW) and Continuous Skip-Gram (CSG) [22,23], and evaluated their adequacy for the objective task described in section 2.1. Following the CBOW algorithm, we randomly initialised m-dimensional embeddings with a Gaussian distribution of mean 0 and standard deviation 1. The text of all samples (including not annotated and annotated, but excluding the 20 samples reserved for validation and final testing; see section 2.2) was then randomly divided into 4.5 million windows of 11 words length each. Each window would contain only words from the same sentence of the central word, using a neutral 'PAD' symbol for positions that spread to other neighbouring sentences. A fully connected single layered network was then created to predict the central word of each window based on the average of all word embeddings appearing within the window. Using this network, embeddings were trained through backpropagation with 0.025 (min alpha 0.0001) learning rate, 5 epochs, and all other parameters set to default values of Word2Vec implementation from the *gensim* library [24]. Separately to CBOW, and following the CSG algorithm, we initialised other set of 100 dimensional embeddings with a Gaussian distribution of mean 0 and standard deviation 1. Text from all not annotated samples was divided into windows in the same manner as done for CBOW. A fully connected single layer network was then trained through with 0.025 (min alpha 0.0001) learning rate and 5 epochs to predict words from the window based on the central embedding. In both cases, the size of the vocabulary consisted on all the words from the embedding and training sets.

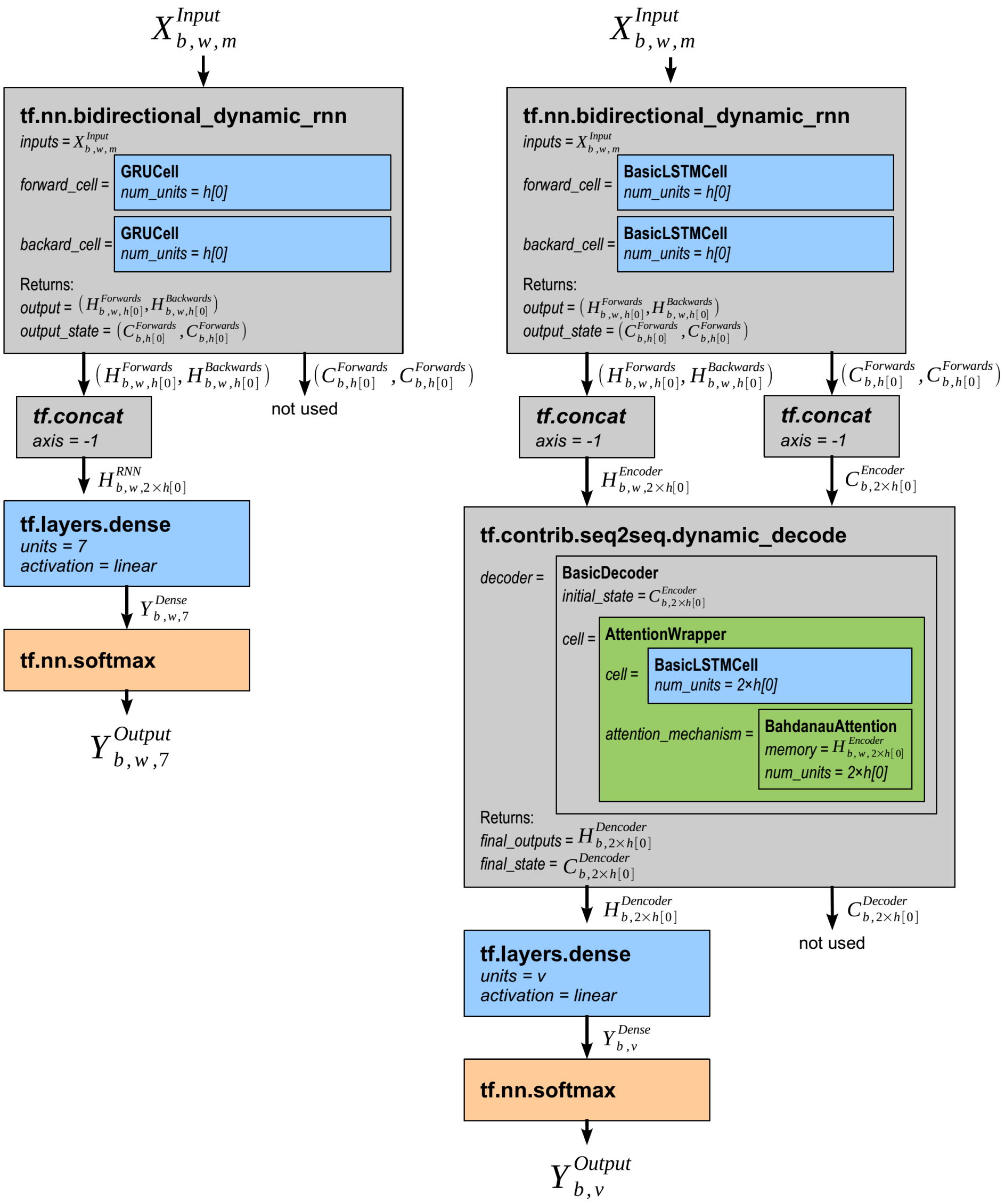

***Figure 2. Architectures for relationship extraction.*** *From left to right, the figure shows the seq2seq and the encoder-decoder RNN architectures. Boxes, colours and fonts have same meaning as in Figure 1.*

### 2.5. Intrinsic evaluation of embeddings

Once created, we intrinsically [25] evaluated the embeddings by calculating their average Euclidean distance, average cosine similarity, and visualising their t-SNE projection. For the first of these, we divided all words into those belonging to each of the target categories (i.e. medication, dosage, mode, frequency, duration, reason; see section 2.1), and those belonging to none. Then we calculated the average Euclidean distance between words of the same class. We followed the same process to calculate the average cosine similarity, but using cosine distance rather than Euclidean distance. Finally, word categories where projected onto a two-dimensional space with t-SNE and then visually inspected to asses class separation [26].

### 2.6. Extrinsic evaluation of embeddings

| Metric | Train | Validation | Test |
|---|---|---|---|
| Num. documents | 238 | 10 | 10 |
| Num. entries | 8387 | 485 | 376 |
| Num. phrases | 21497 | 1329 | 973 |
| Num. tokens | 34718 | 2169 | 1571 |
| Mean entries per document | 35.2 | 48.5 | 37.6 |
| Mean phrases per document | 90.3 | 132.9 | 97.3 |
| Mean tokens per document | 145.9 | 216.9 | 157.1 |
| Mean phrases per entry | 2.6 | 2.7 | 2.6 |
| Mean tokens per entry | 4.1 | 4.5 | 4.2 |
| Mean tokens per phrase | 1.6 | 1.6 | 1.6 |
| Vocabulary of target tokens | 2267 | 442 | 4372 |
| Out of vocabulary tokens | N/A | 48 | 52 |

***Table 2. Document metrics of annotated datasets.*** *The table shows how many documents/entries/phrases/tokens correspond to each of the 3 annotated datasets (train/validation/test) used.*

Besides the three intrinsic evaluation methods described in section 2.5, we also extrinsically evaluated them with a context free classification task [25]. The task consisted on classifying words as either belonging to each of the target classes of the study (i.e. medication, dosage, mode, frequency, duration, reason; see section 2.1) or to none. The task was implemented in the form of a series of binary classifiers, one independently for each target class, and results averaged. The classifier was a feed forwards neural network (FFN) whose input was only the m-dimensional embedding of the to-be-classified word, followed by 'l' densely connected sigmoid layers of 'h' units each, and finally a dense SoftMax layer of 2 units, corresponding with the one-hot representation of the classification objective. Each of the 'h' dense layers was also followed by a dropout operation with proportion 'd' per cent. In the context of this article, we will call this architecture "context free FFN". The training and testing sets were 10000 and 1000 randomly selected words, with 'p'% of them belonging to one of the target classes of the study. The NN was trained with Adam for 'e' epochs, learning rate 'r', using batches of size 'b'. Several values of parameters 'm', 'l', 'h', 'd', 'p', 'e', 'r' and 'b' where tested to prevent using an architecture, dataset or training method that specially favoured either CBOW or CSG.

## 2.7. Term classification

| Field | Train | Validation | Test |
|---|---|---|---|
| Medication | 100% | 100% | 100% |
| Dosage | 49.5% | 56.3% | 50.0% |
| Mode | 37.7% | 40.8% | 37.7% |
| Frequency | 44.8% | 53.4% | 45.4% |
| Duration | 6.1% | 6.0% | 6.1% |
| Reason | 18.3% | 17.5% | 18.1% |

***Table 3. Label metrics of our annotated datasets.*** *The table shows the proportions of entries that contain each of the I2B2 2009 labels.*

The "context free FFN"[1] defined in section 2.6 was also used to obtain a baseline measure of performance on the objective task (section 2.1) with the objective dataset (2.2). In this case we set all free parameters ('m', 'l', 'h', 'd', 'p', 'e', 'r' and 'b') to the values that produced the best performance on the set of words randomly selected in section 2.6.

A second architecture was created by extending the context free FFN into a "context aware FFN"[2]. This consisted on replacing the single word input by the concatenation of the 'w' words existing around the to-be-classified token. Namely, the one-dimensional embedding, which consisted of 'm' real numbers each, were concatenated into a single 1D vector of 'm(1+2w)' real numbers.

A third architecture, partly based on previous work [27], was a "RNN"[3] (recurrent neural network) that sequentially read all words in the target window around the target word. The input to the architecture was one word embedding per time step, fully connected to a LSTM layer of 100 units. The final state of the LSTM layer is fed to a SoftMax function. The NN was trained via Adam algorithm, 0.001 learning rate, 50 batch size, 3 epochs.

## 2.8. Relationship extraction

I2B2 challenge consisted on extracting all medications, dosages, modes, frequencies, durations and reasons as individual terms (see section 2.1), and the architectures of section 2.7 were designed and tested for this objective.

1 In the GitHub repository, this architecture is defined in file 'Model 2 (Feed Forward).ipynb'

2 Defined in file 'Model 3 (Windowed Feed Forward).ipynb'

3 Defined in 'Model 4 (Recurrent).ipynb'

| Field | CBOW | | CSG | |
|---|---|---|---|---|
| | AED | ACS | AED | ACS |
| Medication | 6.53 | 0.23 | 3.61 | 0.53 |
| Dosage | 11.93 | 0.15 | 4.45 | 0.42 |
| Mode | 10.26 | 0.21 | 4.51 | 0.43 |
| Frequency | 14.63 | 0.15 | 4.76 | 0.41 |
| Duration | 17.01 | 0.07 | 4.91 | 0.34 |
| Reason | 12.65 | 0.10 | 4.68 | 0.35 |

***Table 4. Intrinsic evaluation of embeddings.*** *The table shows results of the intrinsic evaluation performed on the embeddings trained either with CBOW or CSG (see section 2.5). AED - Average Euclidean Distance; ACD - Average Cosine Similarity.*

However, in practice, what is of importance is not only the medical terms themselves, but also the relationships between them. Namely, when extracted medical information is used in a subsequent epidemiological analysis, it is of little value to know that a patient took, for example, aspirin, as this patient could have taken the drug in only one occasion, which would have no long-term impact on chronic diseases. What in that example would be of interest is to know whether the patient takes aspirin daily, for how long and with what dosage. Therefore, due to the importance of extracting relationships between medical terms, we also designed and tested a fourth and a fifth architectures specialised on, given a target medication term, extracting its dosage, mode, frequency, duration and reason.

The fourth architecture, which was the first one used for this task, was a sequence to sequence (seq2seq) RNN[4], which sequentially read all words within a 5 rows window around the target medication word, simultaneously outputting word classification at each time step. A bidirectional neural network architecture comprising 100 gated recurrent units (GRU) was initialised with a linear transformation of bag of words (BOW) representation of the target medication for that window. This BOW representation consisted on the sum of all words part of the target medication term (e.g. for the term ‘baby aspirin’, embedding of ‘baby’ plus embedding of ‘aspirin’) concatenated with the medication label, which altogether created a vector of length ‘m’ (size of embeddings) plus 1 (for the medication label). The weights and biases of the linear transformation were learnt during training. Then, the GRU was sequentially fed with the 100-dimensional word embeddings of the sentence, where embeddings were concatenated with an additional real number representing the I2B2 2009 classification of each word, if any (i.e. 1 for medication, 2 dosage, 3 mode, 4 frequency, 5 duration, 6 reason and 0 for ‘none’). In each time step, the state of the GRU was fed to a SoftMax layer of 7 outputs, representing each of the I2B2 2009 term classes (plus a 7$^{th}$ class for ‘none’). The RNN was trained via Adam algorithm, 0.001 learning rate, 50 batch size, 100 epochs.

| Parameter | Context free FFN | Context aware FFN | RNN |
|---|---|---|---|
| m = embeddings dimension | 100 | 100 | 100 |
| w = num window words | - | 5 | 15 |
| l = num layers | 2 | 2 | 1 |
| h = num units per layer | [100, 100] | [500, 100] | [100] |
| d = dropout proportion | 0.0 | 0.0 | 0.0 |
| p = proportion of target words | 0.1 | 0.1 | 0.1 |
| e = num epochs | 5 | 5 | 3 |
| r = learning rate | 0.01 | 0.001 | 0.001 |
| d = decay rate | 0.002 | 0.0 | 0.0 |
| b = batch size | 50 | 50 | 50 |

***Table 5. Used parameters.*** *The table shows the values used for the parameters of each architecture.*

| Parameter | Explored values | Med | Dos | Mod | Fre | Dur | Rea |
|---|---|---|---|---|---|---|---|
| Algorithm | CBOW, CSG | CBOW | CSG | CSG | CBOW | CSG | CBOW |
| Num layers ‘l’ | 1, 2 | 1 | 1 | 1 | 1 | 1 | 1 |
| Activation ‘a’ | tanh, σ, ReLU | σ | σ | σ | σ | σ | σ |
| Dropout ‘d’ | 0.0, 0.2, 0.4 | 0.0 | 0.4 | 0.2 | 0.4 | 0.4 | 0.4 |
| Lean rate ‘r’ | 0.001, 0.01 | 0.01 | 0.01 | 0.01 | 0.01 | 0.01 | 0.01 |

***Table 6. Metaparameters of context free NN.*** *The table shows the best performing set of parameters for each field.*

4 Defined in ‘Model 11 (ELS2S).ipynb’

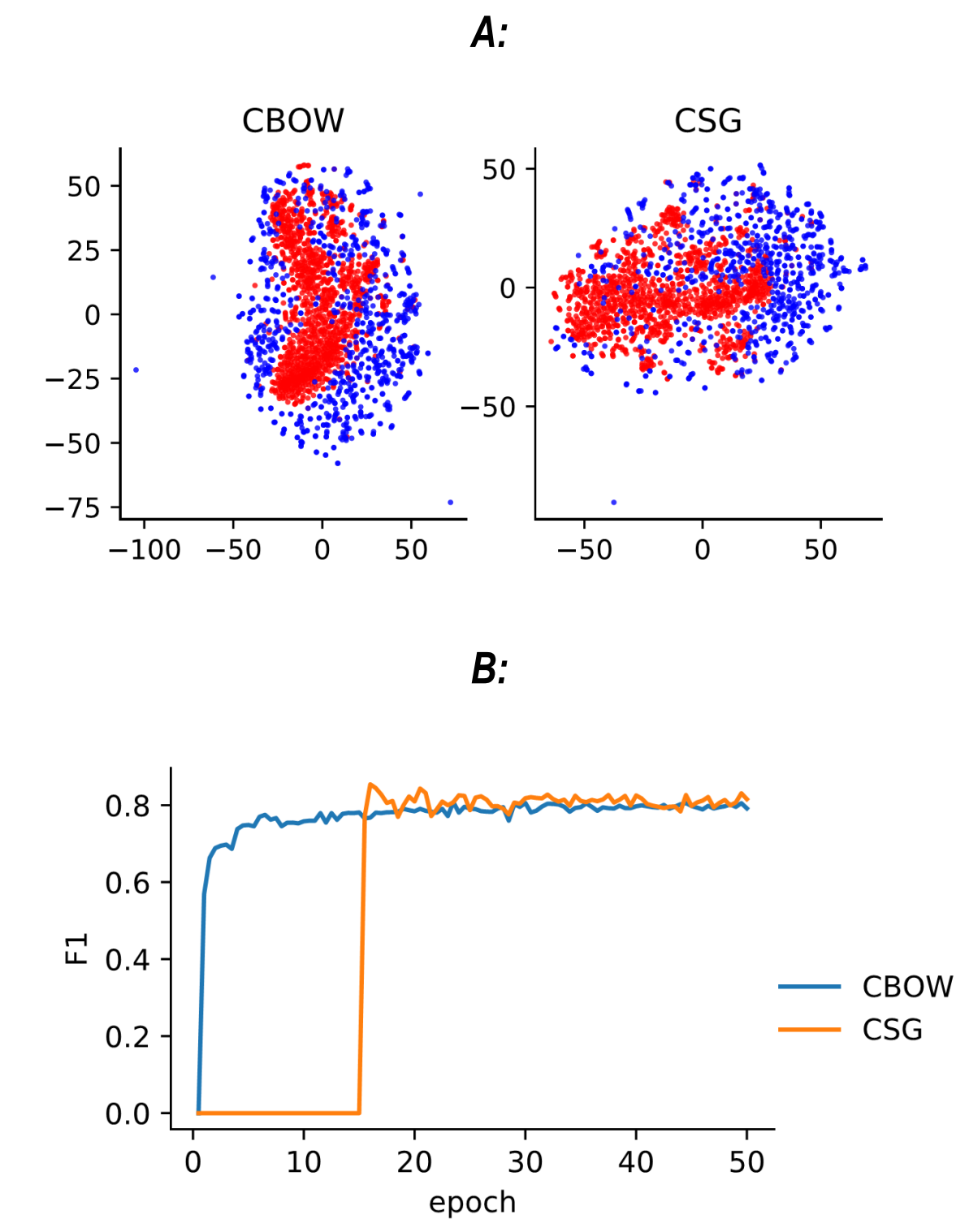

***Figure 3. Evaluation of embeddings. A:*** *Intrinsic evaluation with t-SNE. The figure shows the 2D t-SNE projection of the embeddings calculated with either CBOW (left) or CSG (right). Each point is an embedding, with red corresponding to target categories and blue to other words .**B:** Extrinsic evaluation. The figure shows the F1 score of the context free NN when embeddings are trained using CBOW (blue) or CSG (orange) algorithms.*

| Term | Context free FFN | Context aware FFN | RNN | I2B2 winner |
|---|---|---|---|---|
| Medication | 79.0 | 88.9 | **94.6** | 90.3 |
| Dosage | 71.0 | 91.0 | **93.0** | 90.8 |
| Mode | 95.4 | 92.7 | **96.9** | 89.3 |
| Frequency | 79.8 | 88.5 | **90.9** | 87.7 |
| Duration | 31.7 | 61.9 | **63.0** | 56.0 |
| Reason | 26.5 | 28.1 | 28.4 | **47.0** |

***Table 7. Performance on I2B2 2009 objective task.*** *The table shows F1 scores for each of our three architectures on extracting each of the target terms of I2B2 2009. For comparison, results of the winners of I2B2 challenge are also provided in the last column.*

The fifth and final architecture, which was the second one used for the relationships task, was an encoder-decoder RNN[5], which first read all words within a ±2 row window and then outputted all those words deemed as related to the target medication. A bidirectional LSTM encoder of 128 units was initialised with a BOW representation of the target medication. Then, in the encoding phase, the LSTM read the input window coded as in the seq2seq RNN model described in the previous paragraph. On reaching the end of the window, the final states of the encoder in the forwards and backwards directions are concatenated to form the initial 256-dimensional state of a decoder LSTM. During the decoding phase, this second LSTM received as input the step outputs of the decoder weighted by either Bahdanau [28] or Luong [29] attention mechanism. The decoding LSTM then outputted words until emitting a special <end of output> token. Output words were selected with a SoftMax over the whole vocabulary. The RNN was trained via Adam algorithm with power scheduling rate decay, 0.001 learning rate, 0.00001 decay rate, gradients clipped at value 5, 50 batch size, 100 epochs.

In the case of the latter architecture (encoder-decoder RNN), it should be noted that as the model itself produces words rather than labels, it is impossible to assess its results for field specific Type I errors, so a vocabulary lookup function was used to determine the fields of false positive tokens.

## 3 Results

### 3.1. Text pre-processing

Each document contains a number of entries, which are further divided into sentences and tokens. A number of document metrics count how documents/entries/sentences/tokens correspond to each other. The total number of unique tokens appearing in the unannotated dataset (see Table 2) forms the vocabulary size of our embeddings, which does not include a small number of words of the validation (5) and testing (7) sets. Further labels metrics indicate that pre-annotated terms are evenly distributed across train, validation and testing sets (see Table 2).

### 3.2. Intrinsic evaluation of embeddings

Intrinsic evaluation did not clearly favour one method of constructing embeddings above the other (see Table 4). Average Euclidean distance showed preference for CSG embeddings over CBOW, while average cosine similarity did the opposite. Visual inspection with t-SNE (see Figure

5 Defined in ‘Model 10 (S2S).ipynb’

3) indicated that both methods separated words belonging to target categories (i.e. medication, dosage, mode, frequency, duration, reason; see section 2.1) from the rest, but again without a method clearly outperforming the other. We also explored with embedding sizes of $2^4$ to $2^{10}$ and noticed diminishing improvements in performance at values above $2^7$, ultimately settling at an embedding size of 100.

| Term | seq2seq RNN | encoder-decoder RNN + Bahdanau | encoder-decoder RNN + Luong |
|---|---|---|---|
| Average | 0.824 | 0.806 | 0.811 |
| Medication | 0.897 | 0.851 | 0.876 |
| Dosage | 0.797 | 0.876 | 0.879 |
| Mode | 0.863 | 0.889 | 0.831 |
| Frequency | 0.811 | 0.785 | 0.826 |
| Duration | 0.701 | 0.434 | 0.547 |
| Reason | 0.667 | 0.463 | 0.402 |

***Table 8: Performance on relationship extraction task.*** *The table shows F1 scores for each of our architectures capable of extracting relationships between I2B2 2009 terms and pre-annotated drugs.*

## 3.3. Extrinsic evaluation of embeddings

To further evaluate embeddings, we created a context free FFN whose input was the embedding of a single word and trained it on classifying such words as either belonging to any of the target classes of the study or to none (see section 2.6). The NN meta-parameters that we explored and the values that obtained best performance are in Table 6. One single layer, sigmoid activation functions, dropout at 0.4 and a learning rate of 0.01 obtained in general the best performance. However, no significant difference in F1 was found between CBOW and CSG algorithms (see Figure 3), although CBOW converged earlier and had a more stable final performance.

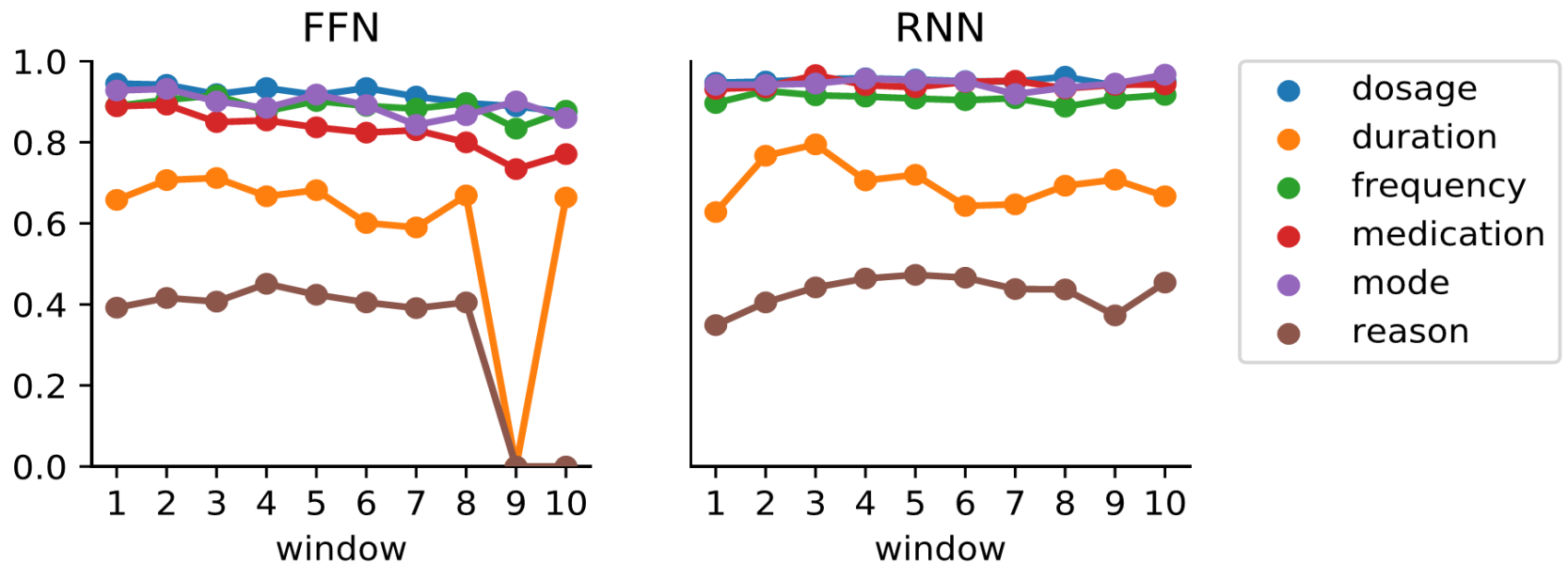

***Figure 4. Effects of window size.*** *The figure shows how F1 varies depending on window size of the context aware FFN (left) and the RNN (right) architectures.*

| | |
|---|---|
| Word embedding input | vancomycin <start> her graft the remainder of the hospital course was unremarkable on, the <num> of july , she was discharged back to the hospital discharge medication vancomycin <num> mg iv q d , ofloxacin <num> mg po bid ( both antiotics to continue for an additional to week course ) , Coumadin with target |
| Word class input | 1 0 6 6 0 0 0 0 0 0 0 0 0 0 0 0 0 0 0 0 0 0 0 0 0 0 0 0 1 2 2 3 4 4 0 1 2 2 3 4 0 0 1 0 0 5 5 5 5 5 5 0 0 1 0 0 |
| Output | Vancomycin <num> mg iv q d for an additional two week course <eos> |

***Figure 5: Term relationship sample.*** *The table shows the two streams of input (word embedding and word class) that both the seq2seq and the encoder-decoder RNNs would receive in this sample. The third row shows the output given by the encoder-decoder after it read this particular example, while the last row shows the ground truth.*

## 3.4. Term classification

Three architectures were trained and tested on the objective task of the original I2B2 2009 challenge. The first architecture was the context free FFN described in section 2.6 with the optimal metaparameter values of section 3.3. The second architecture was a context aware FFN, which extended the previous context free architecture by also reading the '±w' words existing around the to-be-classified token. The third architecture was a LSTM-based RNN capped by a SoftMax that sequentially read the '±w' words existing around the target token. This last architecture outperformed the FNN models in all target terms. Further its performance was above the winner algorithm of I2B2 2009 challenge in all tasks except for extracting 'reason' (see Table 7). Interestingly, context aware FFN preferred small window sizes, while the performance of the RNN was not specially affected by the value of 'w' (see supplementary Figure 4).

## 3.5. Relationship extraction

Beyond the official I2B2 2009 term extraction task, we also created two architectures to identify all terms associated to a given pre-annotated drug (see section 2.8). These were a seq2seq RNN, which simultaneously read the input word by word while outputting word classification, and an encoder-decoder RNN, which first read all input words and then outputted all those related to the pre-annotated drug. The encode-decoder system was trained and tested with two different methods of attention – Bahdanau [28] and Luong [29]. Examples of extractions by these architectures are shown in Figure 5 and the results can be seen in Table 8.

## Discussion:

Architectures based on the artificial neural networks suffer from requiring large amounts of annotated data to be able to perform at state-of-the-art-accuracy. This fact bars them from applications where data is scarce or difficult to access and annotate, such as EHRs. This is the reason why laboratories working with EHRs have traditionally preferred classical methods such as rule-based systems [9,16,30]. In this study we demonstrate that appropriate use of transfer learning and unsupervised learning allow NNs to perform above traditional methods such as those applied earlier [6]. Specifically, fine tuning embeddings to domain specific text (i.e. medical text) and the use of recurrent architectures appeared to produce the highest gains in performance. Interestingly, high dropout rates performed better than low dropout rates only for the terms that were least annotated (see Table 3), even when the most densely annotated terms (e.g. 'medication') were only sampled in 238 documents.

However, our model still did perform poorly for the least annotated categories (e.g. 'reason', see Table 3), where the traditional knowledge-based approaches that won the original challenge achieved better results (see Table 7). The same problem arose for relationship extraction (section 3.5), because each sample was now each record entry (e.g. each record with a word of the category 'medication'), rather than each annotated word (e.g. each word of the category 'medication'), as implied in Figure 5.

Future work could attempt at further improving the performance of NNs in small annotated datasets by transferring learning from unannotated datasets larger than what we used here, and using both within-domain (e.g. medical) and out-of-domain corpora. It is striking that a non-medical expert can learn to recognize reasons for prescribing medications (i.e. our category 'reason') in EHRs after only seen a few examples, while NNs still reach only F1 score of 0.281 even after seeing numerous more examples than a human. To mitigate the problem of learning from scarce data, a few-shot learning approach for medical texts was introduced recently [31]. One of the challenges outlined in above, namely the representation of the worst performing categories, such as 'reasons', could be addressed using fuzzy sets and fuzzy logic due to their ability to capture semantics of vague linguistic constructs due to their capacity [32]. This approach was studied in recent works [33,34], where an adaptive fuzzy control scheme for stochastic non-linear systems was introduced as well as using an efficient representation of high-dimensional ordered data using the path signature from stochastic analysis [35,36,37,38]. Given that knowledge-based methods still outperformed our NN in the 'reason' category (F1=0.47), other avenues could consist on introducing field knowledge into the NN in the form of bias, or in the form of symbolic methods such as dictionaries and gazetteers. Finally, more theoretical work such as the Information Bottleneck [39], the Neural Homology [40], or other theories could allow us to better understand why NNs still need such a large number of samples to learn appropriately, and guide future work on how this problem could be overcome.

## Acknowledgments

AK was supported by MRC Pathfinder Grant Ref: MC-PC-17215. UK CRIS was established and receives continuing funding from the National Institute for Health Research (NIHR) and the Medical Research Council (MRC), including through Dementias Platform UK, and from the NIHR Biomedical Research Centre at Oxford Health National Health Service (NHS) Foundation Trust and the University of Oxford.